\documentclass[runningheads]{llncs}

\usepackage[T2A]{fontenc}

\usepackage{siunitx} 

\usepackage{booktabs}
\usepackage{array} 

\usepackage{amsmath}
\usepackage{amssymb}

\usepackage{enumitem}

\usepackage[ruled,linesnumbered]{algorithm2e}

\usepackage{etoolbox}
\AtBeginEnvironment{algorithm}{\SetArgSty{textrm}}

\usepackage{tikz}
\usetikzlibrary{positioning}

\usepackage{blindtext}
\usepackage{scrextend}
\addtokomafont{labelinglabel}{\sffamily}

\usepackage{float}
\usepackage{graphicx}

\usepackage{subcaption}

\newcommand{\specialcell}[2][c]{%
  \begin{tabular}[#1]{@{}c@{}}#2\end{tabular}}
 
\newcommand*\concat{\mathbin{+}}

\newcommand*\pos{\text{pos}}
\newcommand*\pend{\text{pend}}

\newcommand{\RNum}[1]{\uppercase\expandafter{\romannumeral #1\relax}}

\newcommand\foreign[1]{\emph{#1}}

\newcommand{\bluecomment}[1]{%
    \text{\small{\textcolor{blue}{#1}}}}

\SetKwComment{Comment}{}{}

\SetCommentSty{mycommfont}

\begin{document}
\title{Part of speech and gramset tagging algorithms for unknown words based on morphological dictionaries of the Veps and Karelian languages\thanks{The study was supported by the Russian Foundation for Basic Research, grant 18-012-00117.}}
\titlerunning{POS tagging and gramset algorithms (Veps and Karelian dictionaries)}
%
\author{Andrew Krizhanovsky\inst{1,2}\orcidID{0000-0003-3717-2079} \and
Natalia Krizhanovskaya\inst{1}\orcidID{0000-0002-9948-1910} \and
Irina Novak\inst{3}\orcidID{0000-0002-9436-9460}}
\authorrunning{A. Krizhanovsky et al.}
%
\institute{Institute of Applied Mathematical Research \\ of the Karelian Research Centre of the Russian Academy of Sciences \and  
Petrozavodsk State University \and
Institute of Linguistics, Literature and History \\ of the Karelian Research Centre of the Russian Academy of Sciences, \\ 
Petrozavodsk, Russia,
\email{andrew.krizhanovsky@gmail.com}\\
\url{http://dictorpus.krc.karelia.ru}}
\maketitle              
\begin{abstract}
This research devoted to the low-resource Veps and Karelian languages. 
Algorithms for assigning part of speech tags to words and grammatical properties to words are presented in the article. 
These algorithms use our morphological dictionaries, where the lemma, part of speech and a set of grammatical features (gramset) are known for each word form. 
The algorithms are based on the analogy hypothesis that words with the same suffixes are likely to have the same inflectional models, the same part of speech and gramset.
The accuracy of these algorithms were evaluated and compared. 
66 thousand Karelian 
and 313 thousand Vepsian words 
were used to verify the accuracy of these algorithms.
The special functions were designed to assess the quality of results of the developed algorithms.
\num{86.8}\% of Karelian words 
and \num{92.4}\% of Vepsian words 
were assigned a correct part of speech by the developed algorithm. 
\num{90.7}\% of Karelian words 
and \num{95.3}\% of Vepsian words 
were assigned a correct gramset by our algorithm.
Morphological and semantic tagging of texts, which are closely related and inseparable in our corpus processes, are described in the paper.

\end{abstract}
%
%
\section{Introduction}

Our work is devoted to low-resource languages: Veps and Karelian. These languages belong to the Finno-Ugric languages of the Uralic language family. 
Most Uralic languages still lack full-fledged morphological analyzers and large corpora~\cite{ref_Pirinen_2019}.

In order to avoid this trap the researchers of 
Karelian Research Centre are developing the    
Open corpus of Veps and Karelian languages (VepKar). 
Our corpus contains morphological dictionaries of the Veps language and the three supradialects of the Karelian language: the Karelian Proper, Livvi-Karelian and Ludic Karelian. 
The developed software (corpus manager)\footnote{See
\url{https://github.com/componavt/dictorpus}}
and 
the database, including dictionaries and texts, have open licenses. 

Algorithms for assigning part of speech tags to words 
and grammatical properties to words, 
without taking into account a context, 
using manually built dictionaries, are presented in the article
(see Section~\ref{section:pos_algorithm}).

The proposed technology of evaluation (see the section~\ref{section:experiments}) 
allows to use all 313 thousand Veps and 66 thousand Karelian words to verify the accuracy of the algorithms (Table~\ref{tab:nwords}).
Only a third of Karelian words (28\%) and two-thirds of Veps words (65\%) in the corpus texts are automatically linked to the dictionary entries 
with all word forms (Table~\ref{tab:nwords}).
These words were used in the evaluation of the algorithms.

\begin{table}
\caption{Total number of words in the VepKar corpus and dictionary}\label{tab:nwords}

\begin{tabular}{c c c r} \toprule
Language & \multicolumn{1}{|c|}{\specialcell{The total number\\
                                        of tokens in texts,\\
                                        $10^3$}}
         & \multicolumn{1}{|c|}{\specialcell{N tokens linked to\\
                                    dictionary automatically,\\
                                    $10^3$}}
         & \specialcell{N tokens linked to\\lemmas having\\
                        a complete paradigm,\\
                        $10^3$}\\
\midrule
Veps    & 488     & 400     (82\%) & 313     (65\%)\\
\cmidrule{1-1}
\specialcell{Karelian\\Proper} 
        & 245     & 111     (45\%) & 69     (28\%)\\
\bottomrule
\end{tabular}
\end{table}



Let us describe several works devoted to the development of morphological analyzers for the Veps and Karelian languages.
\begin{itemize}
  \item The Giellatekno language research group is mainly engaged in low-resource languages, the project covers about 50 languages~\cite{ref_Moshagen_2014}. Our project has something in common with the work of Giellatekno in that (1) we work with low-resource languages, (2) we develop software and data with open licenses.
  
  A key role in the Giellatekno infrastructure is given to formal approaches (grammar-based approach) in language technologies. They work with morphology rich languages. Finite-state transducers (FST) are used to analyse and generate the word forms~\cite{ref_Moshagen_2014}.
  
  \item There is a texts and words processing library for the Uralic languages called UralicNLP~\cite{ref_Hamalainen_2019}.
  This Python library provides interface to such Giellatekno tools as FST for processing morphology and constraint grammar for syntax. 
  The UralicNLP library lemmatizes words in 30 Finno-Ugric languages and dialects including the Livvi dialect of the Karelian language (\textit{olo} -- language code).
\end{itemize}


\section{Data organization and text tagging in the VepKar corpus}

Automatic text tagging is an important area of research in corpus linguistics. It is required for our corpus to be a useful resource.

\begin{figure}
\includegraphics[width=0.6\textwidth]{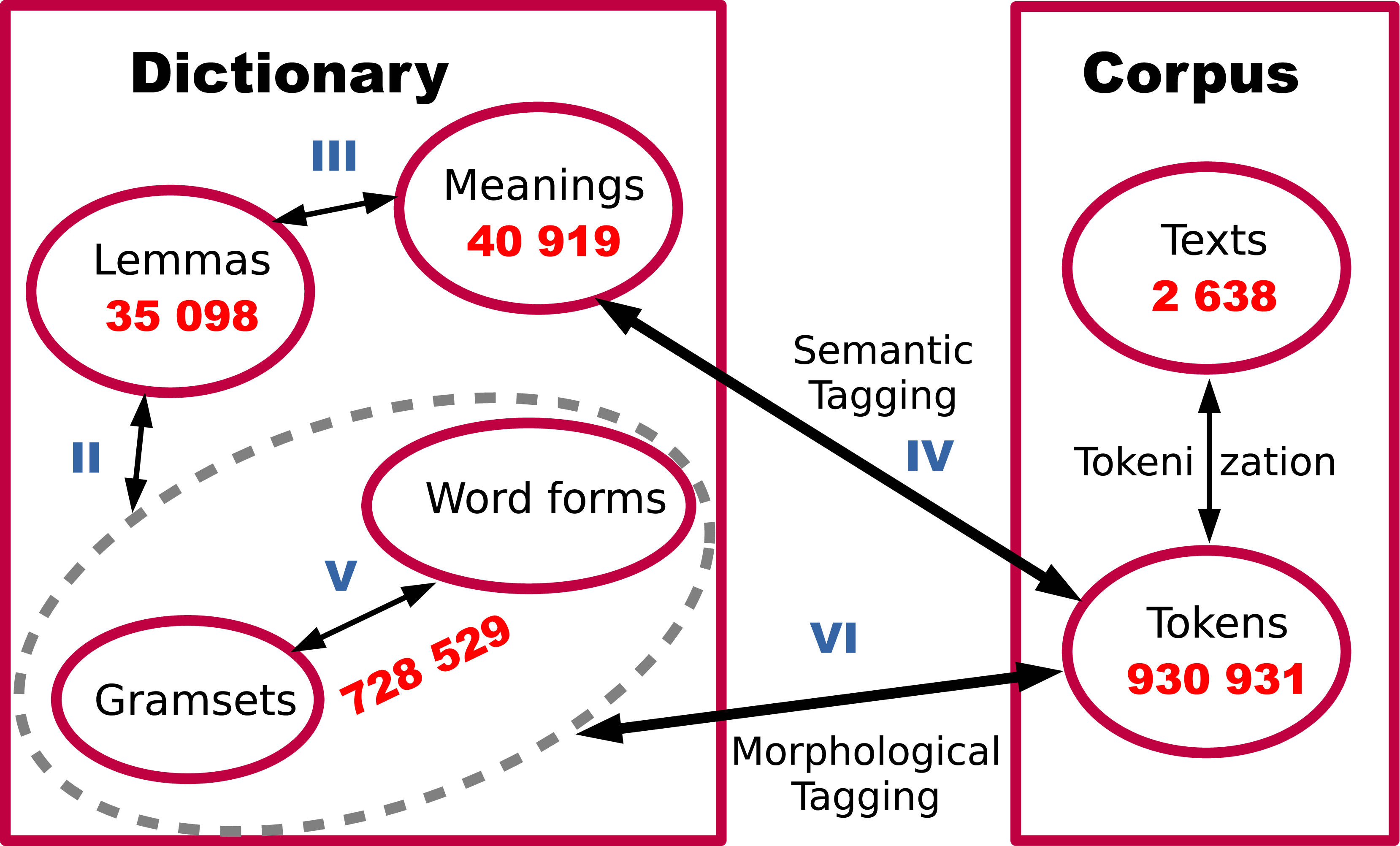}
\caption{Data organization and text tagging in the VepKar corpus.\newline Total values (e.g. number of words, texts) are calculated for all project languages.} \label{fig:corpus:manager:tagging}
\end{figure}

The corpus manager handles the dictionary and the corpus of texts (Fig.~\ref{fig:corpus:manager:tagging}).
The texts are segmented into sentences, 
then sentences are segmented into words (tokens).
The dictionary includes lemmas with related meanings, word forms, and \textbf{sets} of \textbf{gram}matical features (in short -- \textbf{gramsets}).

Text tokens are automatically searched in the dictionary of lemmas and word forms, this is the first stage (\RNum{1}) of the text tagging, it is not presented at Fig.~\ref{fig:corpus:manager:tagging}.
\begin{enumerate}
\item \textbf{Semantic tagging}. 
For the word forms found in the dictionary, the lemmas linked with them are selected (\RNum{2} at Fig.~\ref{fig:corpus:manager:tagging}), then all the meanings of the lemmas are collected (\RNum{3}) and semantic relationships are established between the tokens and the meanings of the lemmas (marked ``not verified'') (\RNum{4}). 

The task of an expert linguist is to check these links and confirm their correctness, either choose the correct link from several possible ones, or manually add a new word form, lemma or meaning.


\begin{tikzpicture}

\node (token)               {tokens (words)};
\node (wf) [right=of token] {word forms};
\node (lemka) [right=of wf] {lemmas};
\node (meaning) [right=of lemka] {meanings};

\draw[<->] (token.east) -- node[above] {\RNum{1}} (wf.west);
\draw[<->] (lemka.east) -- node[above] {\RNum{3}} (meaning.west);
\draw[<->] (wf.east) -- node[above] {\RNum{2}} (lemka.west);
\path[<->,red] (token.south) edge [bend right=13] node[above] {\RNum{4} not verified} (meaning.south);
\end{tikzpicture}


When the editor clicks on the token in the text, then a drop-down list of lemmas with all the meaning will be shown. The editor selects the correct lemma and the meaning (Fig.~\ref{fig:meaning:select}). 





%
\begin{figure}
  \fbox{\includegraphics[width=1\textwidth]{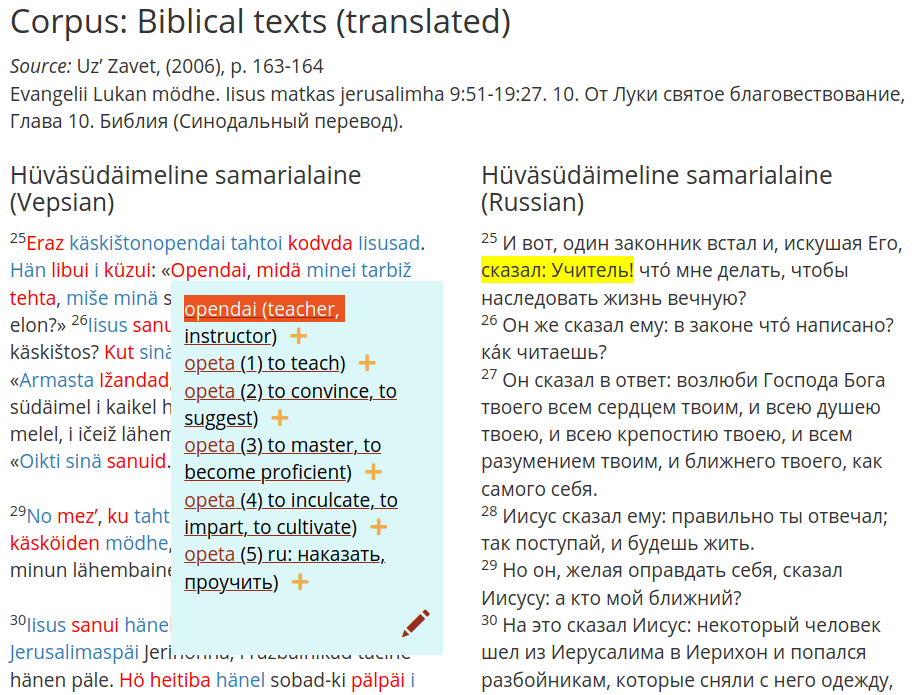}}
%
\caption[Selection the correct lemma and the meaning for token]%
    {Vepsian and Russian parallel Bible 
    translation\textsuperscript{$\ddagger$} 
    in the corpus.
    The editor clicks the word ``Opendai'' in the text, a menu pops up. This menu contains a list of meanings collected automatically for this token, namely: 
    the meaning of the noun ``teacher'' (``opendai'' in Veps) and five meanings of the Veps verb ``opeta''.  
    The noun ``opendai'' and the verb ``opeta'' have the same wordform ``opendai''.
    If the editor selects one of the lemma meanings in the menu (clicks the plus sign), then the token and the correct meaning of the lemma will be connected 
    ({\color{red}\RNum{4} stage is verified}).
    \newline 
}
\line(1,0){82}\newline
\small\textsuperscript{$\ddagger$} See full text online at VepKar: \url{http://dictorpus.krc.karelia.ru/en/corpus/text/494}
    
  \label{fig:meaning:select}
\end{figure}

\item \textbf{Morphological tagging}. 
For the word forms found in the dictionary, the gramsets linked with them are selected (\RNum{5}) and morphological links are established (\RNum{6}) between the tokens and the pairs ``word form -- gramset'' (Fig.~\ref{fig:corpus:manager:tagging}). The expert's task is to choose the right gramset.

\begin{tikzpicture}
\node (token)               {tokens (words)};
\node (wf) [right=of token] {word forms};
\node (gramset) [right=of lemka] {gramsets};

\draw[<->] (token.east) -- node[above] {\RNum{1}} (wf.west);
\draw[<->] (wf.east) -- node[above] {\RNum{5}} (gramset.west);
\path[<->,red] (token.south) edge [bend right=13] node[above] {\RNum{6}} (gramset.south);
\end{tikzpicture}

\end{enumerate}


\section{Corpus tagging peculiarities}\label{section:corpus_peculiarities}
In this section, we describe why the word forms with white spaces and analytical forms are not taken into account in the search algorithm described below.
Analytical form is the compound form consisting of auxiliary words and the main word.

The ultimate goal of our work is the morphological markup of the text, previously tokenized into words by white spaces and non-alphabetic characters 
(for example, brackets, punctuation, numbers).
Therefore, analytical forms do not have markup in the texts.

Although we store complete paradigms in the dictionary, including analytical forms, such forms do not used in the analysis of the text, because each individual word is analyzed in the text, not a group of words.

For example, we take the Karelian verb ``pageta'' (leave, run away).
In the dictionary not only the negative form of indicative, presence, first-person singular ``en pagene'' is stored, but also connegative (a word form used in negative clauses) of the indicative, presence ``pagene'', which is involved in the construction of five of the six forms of indicative, presence.
Thus, in the text the word ‘en’ (auxiliary verb ‘ei’, indicative, first-person singular) and ‘pagene’ (verb ‘pageta’, connegative of indicative, presence) are separately marked.

\section{Part of speech and gramset search by analogy algorithms}\label{section:pos_algorithm}

The proposed algorithms operate on data from a morphological dictionary. 
%
The algorithms are based on the analogy hypothesis that words with the \emph{same suffixes} are likely to have the same inflectional models and the same sets of grammatical information (part of speech, number, case, tense, etc.). 
The \emph{suffix} here is a final segment of a string of characters.

%
%
Let the hypothesis be true, in that case, if the suffixes of new words coincide with the suffixes of dictionary words, then part of the speech and other grammatical features of the new words will coincide with the dictionary words.
%
%
It should be noted that 
the length of the suffixes is unpredictable and can be various for different pairs of words~\cite[p.~53]{ref_Belonogov_2004}.

%
The POSGuess and GramGuess algorithms described below use the concept of ``suffix'' (Fig.~\ref{fig:kezaman_substring}), the GramPseudoGuess algorithm uses the concept ``pseudo-ending'' (Fig.~\ref{fig:huukkua_substring}).

\subsection{The POSGuess algorithm for part of speech tagging with a~suffix}

Given the set of words $W$, for each word in this set the part of speech is known. 
The algorithm~\ref{alg:POSGuess} finds a part of speech $\pos_u$ for a given word~$u$ using this set. 

\begin{algorithm}
    \caption{Part of speech search by a suffix (POSGuess)}
    \label{alg:POSGuess}
\DontPrintSemicolon
\SetAlgoLined
    \KwData{
        $P$ -- a set of part of speech (POS),\hfill \break
        $W = \{ w \; | \; \exists \pos_w \in P \}$ -- a set of words, POS is known for each word,\hfill \break
        $u \notin W$ -- the word with unknown POS,\hfill \break
        $len(u)$ -- the length (in characters) of the string $u$.
       }
    \KwResult{\[
        u_z : 
        \begin{cases}
            %
            len (u_z) \xrightarrow[ z=2,\ldots,len(u) ]{} \max,\;
            \Comment{//\,Longest suffix}\\
            \exists w \in W : w = w_{\text{prefix}} \concat u_z\;
            \Comment{//\,Concatenation of strings}
        \end{cases}
        \]
    \begin{align*}
\text{Counter} & \left[ \pos^k \right] = c^k, \; k = \overline{1,m}, \; \text{where :}\\
        & c^k \in \mathbb{N}, \; c^1 \geq c^2 \geq \ldots \geq c^m,\\
        & \exists w^k_i \in W : 
            w^k_i = {w_{\text{prefix}}}^k_i \concat u_z \Rightarrow
            c^k = \vert \pos^k_{w^k_i} \vert, \\
        &    
            i = \overline{1,c^k}, \\
        & \forall i : \pos^k_{w^k_i} = \pos^k \in P, \;\;\;\; 
           a \ne b \Leftrightarrow \pos^a \ne \pos^b\\
        & 
        m \; \text{-- the number of different POS of found words} \; w^k_i
    \end{align*}
    } 
    \BlankLine
    
    $z$ = 2 \Comment{// The position in the string $u$}
    $z_{found} = \text{FALSE}$ \;
    \BlankLine
    \While{ $z \leq len(u)$ and $\neg z_{found}$ }{
        \BlankLine
        \Comment{// The suffix of the word $u$ from $z$-th character}
        $u_z = \text{ substr } (u, z) \;$
        \BlankLine
        \ForEach{$w \in W$}{
            \Comment{// If the word $w$ has the suffix $u_z$ (regular expression)
            }
            \If{$w =\sim \textrm{m}/u_z\$/$}{
                $\text{Counter} \left[ \pos_w \right] ++$\\
                $z_{found} = \text{TRUE}$ \Comment{// Only POS of words with this $u_z$ suffix will be counted. The next "while" loop will break, so the shorter suffix $u_{z+1}$ will be omitted.}
            }
        }
    z = z + 1
    }
    \BlankLine
    \Comment{// Sort the array in descending order, according to the value}
    \verb|arsort|( $\text{Counter} \left[ \; \right]$ )
\end{algorithm}

\newpage

\begin{figure}
\includegraphics[width=0.7\textwidth]{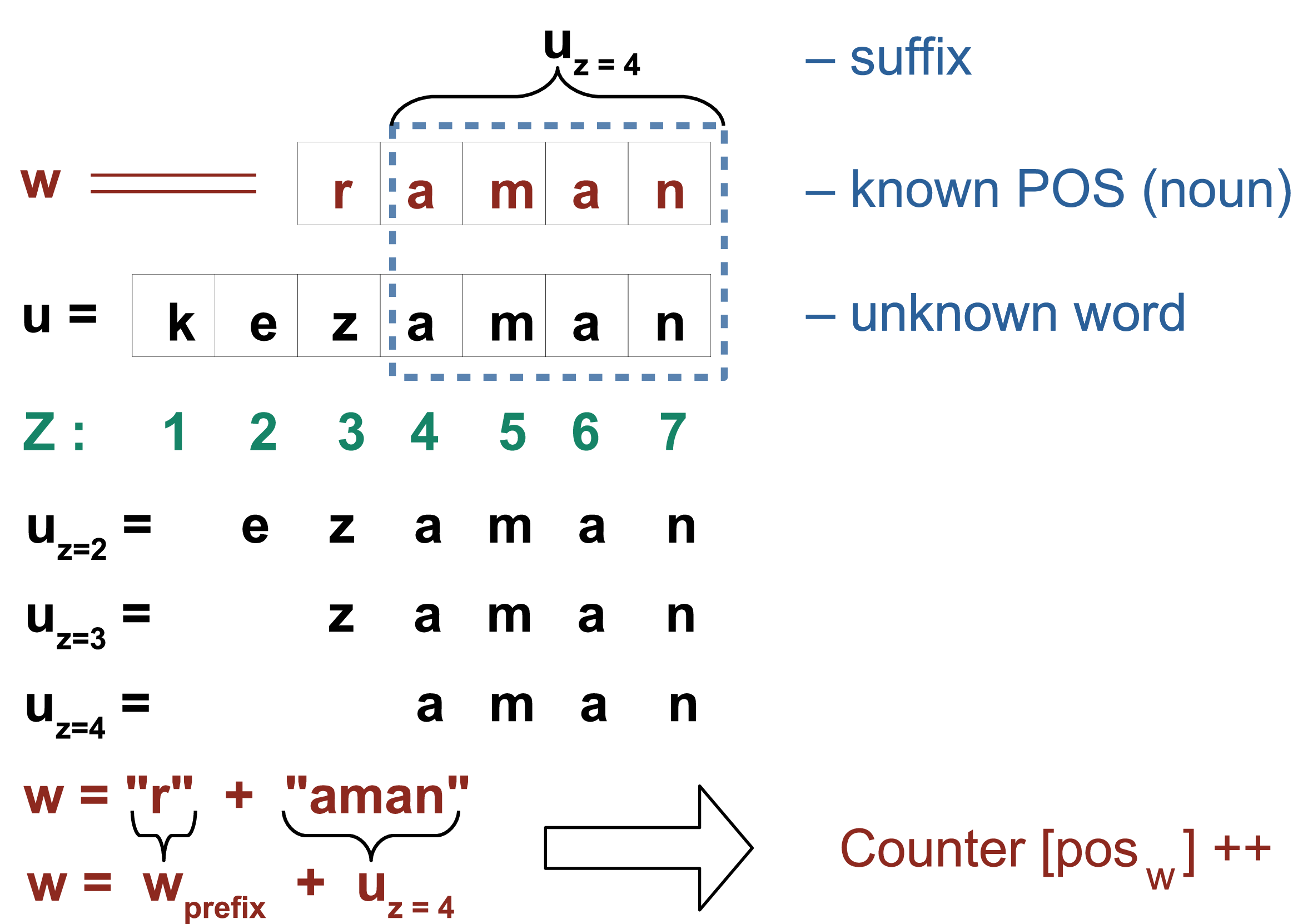}
\caption{Veps nouns in the genitive case ``kezaman'' (``kezama'' means ``melted ground'') and ``raman'' (``rama'' means ``frame''). The word $u$ 
with an unknown part of speech is ``kezaman''. The word $w$ from the dictionary with the known POS is ``raman''. 
They share the common suffix $u_z$, which is ``aman''.} \label{fig:kezaman_substring}
\end{figure}

In Algorithm~\ref{alg:POSGuess} 
we look for in the set $W$ (line~5) the words 
which have the same suffix $u_z$ as the unknown word $u$. 
Firstly, we are searching for the longest substring of $u$, 
that starts at index z. 
The first substring $u_{z=2}$ will start at the second character (line 1 in Algorithm~\ref{alg:POSGuess}), since $u_{z=1} = u$ is the whole string 
(Fig.~\ref{fig:kezaman_substring}). 

Then we increment the value $z$ decreasing the length of the substring $u_z$ in the loop, 
while the substring $u_z$ has non-zero length, $z \leq len(u)$.
If there are words in $W$ with the same suffix, then we count the number of similar words for each part of the speech and stop the search.

\hfill \break


%
The Fig.~\ref{fig:kezaman_substring} shows the idea of the algorithm~\ref{alg:POSGuess}: for a new word ($kezaman$), we look for a word form in the dictionary ($raman$) with the same suffix ($aman$).


We begin to search in the dictionary for word forms with the suffix $u_{z=2}$. If we have not find any words, then we are looking for $u_{z=3}$ and so on.
The longest suffix $u_{z=4}$=``aman'' with $z=4$ is found.

%
Then we find all words with the suffix $u_{z=4}$ and count 
how many of such words are nouns, verbs, adjectives and so on.
The result is written to the array $Counter[\,]$. In Fig.~\ref{fig:kezaman_substring} the \emph{noun} ``raman'' was found, therefore we increment the value of $Counter[noun]$.

\subsection{The GramGuess algorithm for gramset tagging with a suffix}

The GramGuess algorithm is exactly the same as the POSGuess algorithm, 
except that it is needed to search a subset of gramsets instead of parts of speech.
That is in the set $W$ the gramset is known for each word. 
The gramset is a set of morphological tags (number, case, tense, etc.).

\subsection{The GramPseudoGuess algorithm for gramset tagging with a pseudo-ending}

%
Let us explain the ``pseudo-ending'' used in the algorithm GramPseudoGuess.

\begin{figure}
\includegraphics[width=0.5\textwidth]{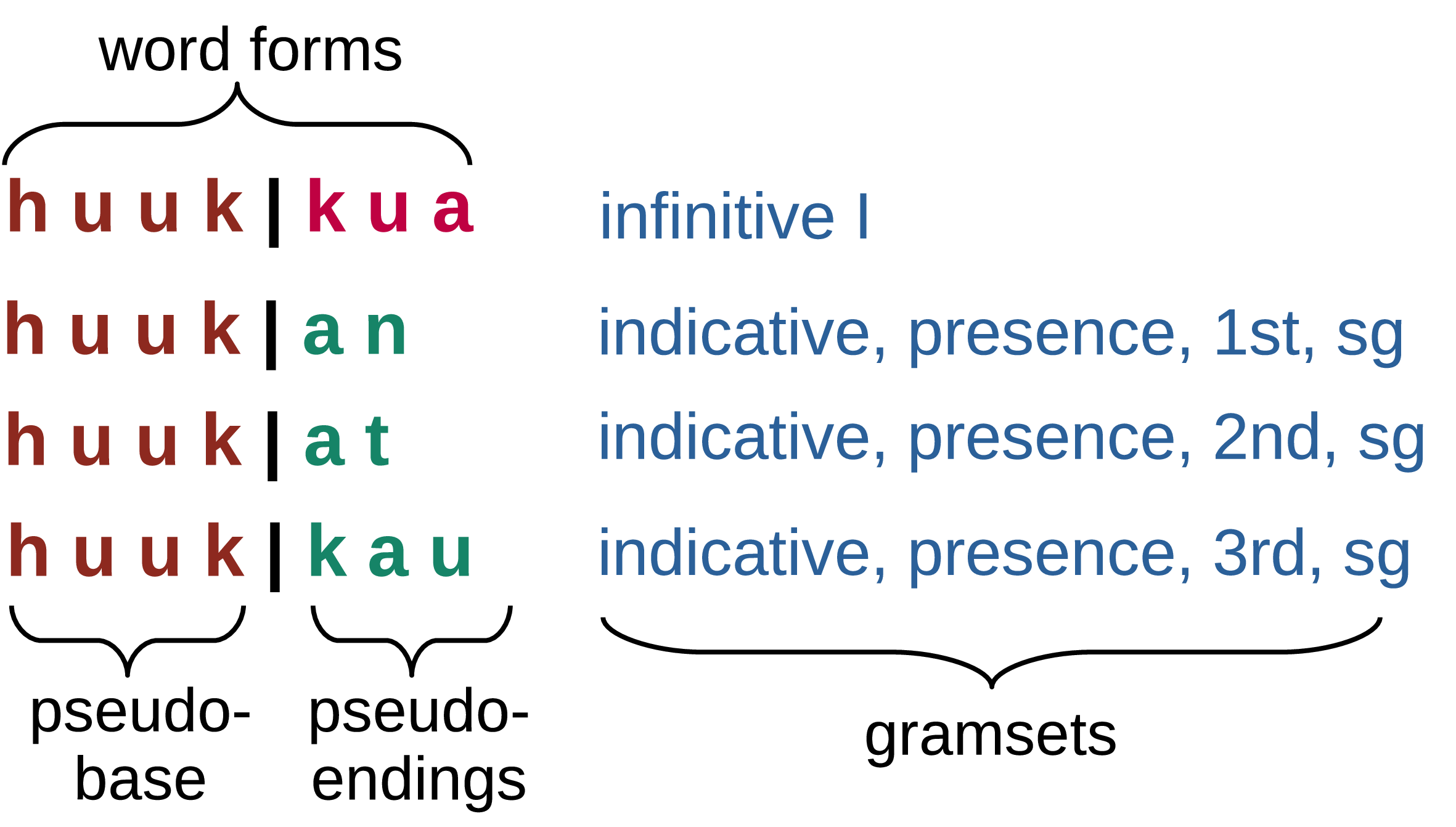}
\caption{Wordforms of the Karelian verb ``huukkua'' (it means ``to call out'', ``to holler'', ``to halloo''). All word forms have the same pseudo-base and different pseudo-endings for different set of grammatical attributes (gramsets).} \label{fig:huukkua_substring}
\end{figure}

All word forms of one lemma share a common invariant substring. This substring is a \emph{pseudo-base} of the word (Fig.~\ref{fig:huukkua_substring}). 
Here the pseudo-base is placed at the start of a word, it suits for the Veps and Karelian languages.
For example, in Fig.~\ref{fig:huukkua_substring} the invariant substring ``huuk'' is the pseudo-base for all word forms of the lemma ``huukkua''. 
The Karelian verb ``huukkua'' means ``to call out'', ``to holler'', ``to halloo''.



Given the set of words $W$, for each word in this set a gramset and a pseudo-ending are known. 
The algorithm~\ref{alg:POSPseudoGuess} finds a gramset $g_u$ for a given word~$u$ using this set.

\begin{algorithm}
    \caption{Gramset search by a pseudo-ending (GramPseudoGuess)}
    \label{alg:POSPseudoGuess}
\DontPrintSemicolon
\SetAlgoLined
    \KwData{
        $G$ -- a set of gramsets,\hfill \break
        $W = \{ w \; | \; \exists \,g_w \in G, 
                \exists \,\pend_w : w = w_{\text{prefix}} \concat \pend_w \}$ 
                -- a set of words, where gramset and pseudo-ending (pend) are known for each word,  \hfill \break
        $u \notin W$ -- the word with unknown gramset,\hfill \break
        $len(u)$ -- the length (in characters) of the string $u$.
       }
    \KwResult{\[
        u_z : 
        \begin{cases}
            %
            len (u_z) \xrightarrow[ z=2,\ldots,len(u) ]{} \max,\;
            \Comment{//\,Longest substring}\\
            \exists w \in W : \pend_w = u_z\;
        \end{cases}
        \]
    \begin{align*}
\text{Counter} & \left[ g^k \right] = c^k, \; k = \overline{1,m}, \; \text{where :}\\
        & c^k \in \mathbb{N}, \; c^1 \geq c^2 \geq \ldots \geq c^m,\\
        & \exists w^k_i \in W : 
            \pend_{w^k_i} = u_z \Rightarrow
            c^k = \vert g^k_{w^k_i} \vert, \\
        &    
            i = \overline{1,c^k}, \\
        & \forall i : g^k_{w^k_i} = g^k \in G, \;\;\;\; 
           a \ne b \Leftrightarrow g^a \ne g^b\\
        & 
        m \; \text{-- the number of different gramsets of found words} \; w^k_i
    \end{align*}
    } 
    \BlankLine
    
    $z$ = 2 \Comment{// The position in the string $u$}
    $z_{found} = \text{FALSE}$ \;
    \BlankLine
    \While{ $z \leq len(u)$ and $\neg z_{found}$ }{
        \BlankLine
        \Comment{// The substring of the word $u$ from $z$-th character}
        $u_z = \text{ substr } (u, z) \;$
        \BlankLine
        \ForEach{$w \in W$}{
            \Comment{// If the word $w$ has the pseudo-ending $u_z$}
            \If{$\pend_w == u_z$}{
                $\text{Counter} \left[ g_w \right] ++$\\
                $z_{found} = \text{TRUE}$ \Comment{// Only gramsets of words with the pseudo-ending $u_z$ will be counted. The next "while" loop will break, so the shorter $u_{z+1}$ will be omitted.}
            }
        }
    z = z + 1
    }
    \BlankLine
    \Comment{// Sort the array in descending order, according to the value}
    \verb|arsort|( $\text{Counter} \left[ \; \right]$ )
\end{algorithm}

In Algorithm~\ref{alg:POSPseudoGuess} 
we look for in the set $W$ (line~5) the words 
which have the same pseudo-ending $u_z$ as the unknown word $u$. 
Firstly, we are searching for the longest substring of $u$, 
that starts at index z. 

Then we increment the value $z$ decreasing the length of the substring $u_z$ in the loop, 
while the substring $u_z$ has non-zero length, $z \leq len(u)$.
If there are words in $W$ with the same pseudo-ending, then we count the number of similar words for each gramset and stop the search.


\section{Experiments}\label{section:experiments}

\subsection{Data preparation}
Lemmas and word forms from our morphological dictionary were gathered to \emph{one set} as a search space of part of speech tagging algorithm.
This set contains unique pairs ``word -- part of speech''.

In order to search a gramset, we form the set consisting of (1) lemmas without inflected forms (for example, adverbs, prepositions) and (2) inflected forms (for example, nouns, verbs).
This set contains unique pairs ``word -- gramset''. For lemmas without inflected forms the gramset is empty.

We put on constraints for the words in both sets: strings must consist of more than two characters and must not contain whitespace.
That is, analytical forms and compound phrases have been excluded from the sets (see section~\ref{section:corpus_peculiarities}).

\subsection{Part of speech search by a suffix (POSGuess algorithm)}

For the evaluation of the quality of results of the searching algorithm POSGuess the following function $\text{eval}(\pos^u)$ was proposed:
\begin{equation}\label{eq:metric-eval}
\text{eval} \left( 
    \begin{array}{@{}l@{\thinspace}l}
       \pos^u, \\
       \text{Counter} & \left[ \pos^k \right] \rightarrow c^k,\\
                      & \forall k = \overline{1,m} \\
    \end{array}
    \right)     = 
        \begin{cases}
            \multicolumn{2}{l}{\bluecomment{The array Counter[\,] 
                  do not contain the correct 
                  $\pos^u$.}}\\
            0, & \pos^u \ne \pos^k, \forall k = \overline{1,m},\\[3mm]
            \multicolumn{2}{l}{\bluecomment{\specialcell{First several POS in the array can have \\ 
                  the same maximum frequency $c^1$, 
                  one of this POS is $\pos^u$.}}}\\
            1, & \pos^u \in \{ \left[ \pos^1, \ldots, \pos^j \right] : 
                               c^1=c^2=\ldots=c^j, j \leq m \},\\[4mm]
            \multicolumn{2}{l}{\frac{c^k}{ \sum_{k=1}^{m}c^k }, \; 
                \exists k : \pos^k = \pos^u, \;
                            c^k < c^1
                }
        \end{cases}
\end{equation}

%
This function~(\ref{eq:metric-eval}) evaluates the result of the POSGuess algorithm against the correct part of speech $\pos^u$.
%
%
The POSGuess algorithm counts the number of words similar to the word $u$ 
separately for each part of speech 
and stores the result in the Counter array.

The Counter array is sorted in descending order, according to the value. The first element in the array is a part of speech with maximum number of words similar to the unknown word $u$.

%
\num{71 091} ``word -- part of speech'' pairs for the Proper Karelian supradialect and \num{399 260} ``word -- part of speech'' pairs for the Veps language have been used in the experiments to evaluate algorithms.

During the experiments, two Karelian words were found, for which there were no suffix matches in the dictionary. They are the word ``cap'' (English: snap; Russian: \foreign{цап}) and the word ``štob'' (English: in order to; Russian: \foreign{чтобы}).
That is, there were no Karelian words with the endings -p and -b.
This could be explained by the fact that these two words migrated from Russian to Karelian language.

%

\begin{figure}[H]
\includegraphics[width=1\textwidth]{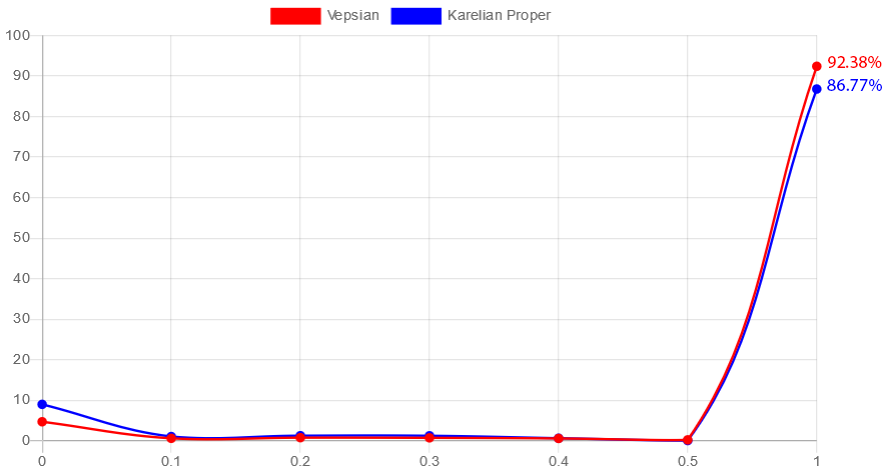}
\caption{
The proportion of Vepsian (red curve) and Karelian (blue curve) words with correct ($x=1$) and wrong ($x=0$) part of speech assignment by the POSGuess algorithm 
with the formula~(\ref{eq:metric-eval}).
} \label{fig:search_POS}
\end{figure}

Figure~\ref{fig:search_POS} shows the proportion of Veps and Karelian words with correct and wrong part of speech assignment by the POSGuess algorithm. 
Values along the X axis are the values of the function $\text{eval}(\pos^u)$, 
see the formula~(\ref{eq:metric-eval}). 
This function for evaluating the part of speech assignment takes the following values:

\begin{labeling}{eval-formula}
\item [0] \num{4.7}\% of Vepsian words and 9\% of Karelian words ($x = 0$ in Fig.~\ref{fig:search_POS}) were assigned the wrong part of speech. 
That is, there is no correct part of speech in the result array $Counter[\,]$ in the \mbox{POSGuess} algorithm.
This is the first line in the formula~(\ref{eq:metric-eval}).
\hfill \break

\item [\num{0.1} -- \num{0.5}] 
\num{2.92}\% of Vepsian words and \num{4.23}\% of Karelian words ($x \in [0.1 ; 0.5]$ in Fig.~\ref{fig:search_POS}) were assigned the partially correct POS tags.
That is, the array $Counter[\,]$ contains the correct part of speech, but it is not at the beginning of the array. 
This is the last line in the formula~(\ref{eq:metric-eval}).
\hfill \break

\item [1] 
\num{92.38}\% of Vepsian words and \num{86.77}\% of Karelian words ($x = 1$ in Fig.~\ref{fig:search_POS}) were assigned the correct part of speech. 
The array $Counter[\,]$ contains the correct part of speech at the beginning of the array. 
\end{labeling}

%
Figure~\ref{fig:search_POS} shows the evaluation of the results of the POSGuess algorithm for all parts of speech together.
%
Table~\ref{tab:POS:quantity:Veps} (Veps) and Table~\ref{tab:POS:quantity:krl} (Karelian) show the evaluation of the same results of the POSGuess algorithm, 
but they are presented for each part of speech separately.


\begin{table}
\caption{Number of Vepsian words of different parts of speech 
used in the experiment. 
The evaluation of results found by POSGuess algorithm by the formula~(\ref{eq:metric-eval}) 
and fraction of results in percent, where 
the column \textit{0} means the fraction of words with incorrectly found POS, \textit{1} -- the fraction of words with correct POS in the top 
of the list created by the algorithm.}\label{tab:POS:quantity:Veps}

\setlength{\tabcolsep}{8pt}

\begin{tabular}{ l r r r r r r r >{\bfseries}r } \toprule
     \multicolumn{2}{c}{\large Veps} &
     \multicolumn{7}{c}{\specialcell{Fraction 
     of not guessed (column 0),\\
     partly guessed (0.1--0.5) and guessed (1) POS, \%}} \\ \cmidrule(r){3-9}
 POS & Words & 0 & 0.1 & 0.2 & 0.3 & 0.4 & 0.5 & \multicolumn{1}{c}{1}\\ \midrule
Verb & \num{93 047} & 2.12 & 0.52 & 0.55 & 0.47 & 0.36 & 0.01 & 95.97\\
Noun & \num{240 513} & 2.88 & 0.3 & 0.67 & 0.6 & 0.42 & 0.24 & 94.89\\
Adjective & \num{61 845} & 12.45 & 1.62 & 1.44 & 1.53 & 1.58 & 0.51 & 80.87\\
Pronoun & \num{1 244} & 46.54 & 8.12 & 0.56 & 0.64 & 0 & 0 & 44.13\\
Numeral & \num{1 200} & 44 & 6.25 & 2.33 & 0.67 & 0.33 & 0 & 46.42\\
Adverb & 650 & 64.92 & 3.08 & 2.46 & 1.23 & 0.46 & 0 & 27.85\\
\bottomrule
\end{tabular}
\end{table}


\begin{table}
\caption{Number of Karelian words of different parts of speech used in the experiment.}\label{tab:POS:quantity:krl}

\setlength{\tabcolsep}{8pt}

\begin{tabular}{ l r r r r r r r >{\bfseries}r } \toprule
     \multicolumn{2}{c}{\large  Karelian} &
     \multicolumn{7}{c}{\specialcell{Fraction 
     of not guessed (column 0),\\
     partly guessed (0.1--0.5) and guessed (1) POS, \%}} \\ \cmidrule(r){3-9}
 POS & Words & 0 & 0.1 & 0.2 & 0.3 & 0.4 & 0.5 & \multicolumn{1}{c}{1}\\ \midrule
Verb & \num{26 033} & 3.26 & 0.5 & 0.74 & 0.6 & 0.23 & 0.01 & 94.67\\
Noun & \num{36 908} & 5.47 & 0.38 & 1.13 & 1.08 & 0.52 & 0.04 & 91.38\\
Adjective & \num{6 596} & 35.81 & 6.66 & 4.15 & 4.56 & 2.73 & 0.38 & 45.71\\
Pronoun & 610 & 81.64 & 2.13 & 0.66 & 3.11 & 2.3 & 0 & 10.16\\
Numeral & 582 & 65.81 & 1.72 & 1.03 & 0.17 & 1.03 & 0 & 30.24\\
Adverb & 235 & 68.51 & 3.4 & 2.98 & 2.13 & 0 & 0 & 22.98\\
\bottomrule
\end{tabular}
\end{table}

\begin{figure}
\includegraphics[width=1\textwidth]{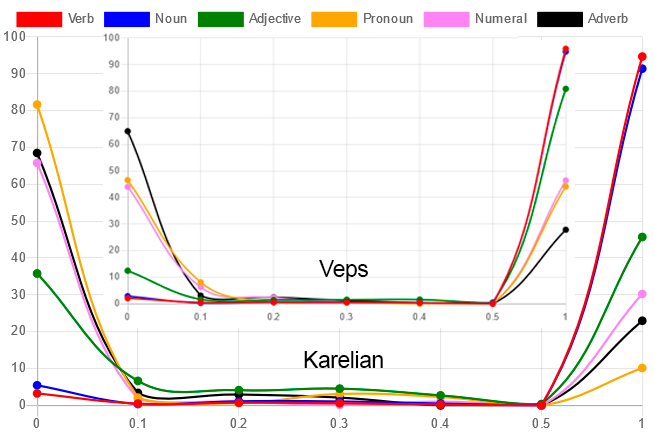}
\caption{Number of Vepsian and Karelian words of different parts of speech used in the experiment.} \label{fig:POS:distribution}
\end{figure}

\subsection{Gramset search by a suffix (GramGuess algorithm) and by a pseudo-ending (GramPseudoGuess algorithm)}

\num{73 395} ``word -- gramset'' pairs for the Karelian Proper supradialect and \num{452 790} ``word -- gramset'' pairs for the Veps language have been used in the experiments to evaluate GramGuess and GramPseudoGuess algorithms.

A list of gramsets was searched for each word. The list was ordered by the number of similar words having the same gramset.

For the evaluation of the quality of results of the searching algorithms the following function $\text{eval}(g^u)$ has been proposed:
\begin{equation}\label{eq:metric-eval-gram}
\text{eval} \left( 
    \begin{array}{@{}l@{\thinspace}l}
       g^u, \\
       \text{Counter} & \left[ g^k \right] \rightarrow c^k,\\
                      & \forall k = \overline{1,m} \\
    \end{array}
    \right)     = 
        \begin{cases}
            \multicolumn{2}{l}{\bluecomment{The array Counter 
                  do not contain the correct gramset 
                  $g^u$.}}\\
            0, & g^u \ne g^k, \forall k = \overline{1,m},\\[3mm]
            \multicolumn{2}{l}{\bluecomment{\specialcell{First several gramsets in the array can have \\ 
                  the same maximum frequency $c^1$, 
                  one of these gramsets is $g^u$.}}}\\
            1, & g^u \in \{ \left[ g^1, \ldots, g^j \right] : 
                               c^1=c^2=\ldots=c^j, j \leq m \},\\[4mm]
            \multicolumn{2}{l}{\frac{c^k}{ \sum_{k=1}^{m}c^k }, \; 
                \exists \, k : g^k = g^u, \;
                            c^k < c^1
                }
        \end{cases}
\end{equation}

%
This function~(\ref{eq:metric-eval-gram}) evaluates the results of the GramGuess and GramPseudoGuess algorithms against the correct gramset $g^u$.

\begin{table}
\caption{Evaluations of results of gramsets search for Vepsian and Karelian by GramGuess and GramPseudoGuess algorithms.
}\label{tab:Gram:quantity}

\setlength{\tabcolsep}{8pt}

\begin{tabular}{ c c c c c} \toprule
     & \multicolumn{2}{c}{GramGuess}& \multicolumn{2}{c}{GramPseudoGuess}\\ \cmidrule(r){2-5}
                                                
 Evaluation & Veps & Karelian & Veps & Karelian\\ \midrule
0 & 2.53 & 5.72 & 7.9 & 9.23\\
0.1 & 0.53 & 0.83 & 1.04 & 1.57\\
0.2 & 0.71 & 1.16 & 1.24 & 1.37\\
0.3 & 0.64 & 0.89 & 2.68 & 1.36\\
0.4 & 0.2 &	0.56 & 0.14 & 0.68\\
0.5 & 0.11 & 0.09 & 0.83 & 0.43\\
1 &	\bf{95.29} & \bf{90.74} & \bf{86.17} & \bf{85.36}\\
\bottomrule
\end{tabular}
\end{table}

%
The table 4 shows that the GramGuess algorithm gives the better results than the GramPseudoGuess algorithm, namely: 
\begin{labeling}{eval-formula}
\item [Karelian] 
90.7\% of Karelian words were assigned a correct gramset by GramGuess algorithm versus 85.4\% by GramPseudoGuess algorithm;
\hfill \break

\item [Veps] 
95.3\% of Vepsian words were assigned a correct gramset by GramGuess algorithm versus 85.4\% by GramPseudoGuess algorithm.
\end{labeling}

It may be suggested by the fact that suffixes are longer than pseudo-endings. 
%
In addition, the GramPseudoGuess algorithm is not suitable for a part of speech without inflectional forms.

\section{Morphological analysis results}

\begin{figure}[H]
\centering
\begin{subfigure}{.5\textwidth}
  \centering
  \includegraphics[width=0.95\textwidth]{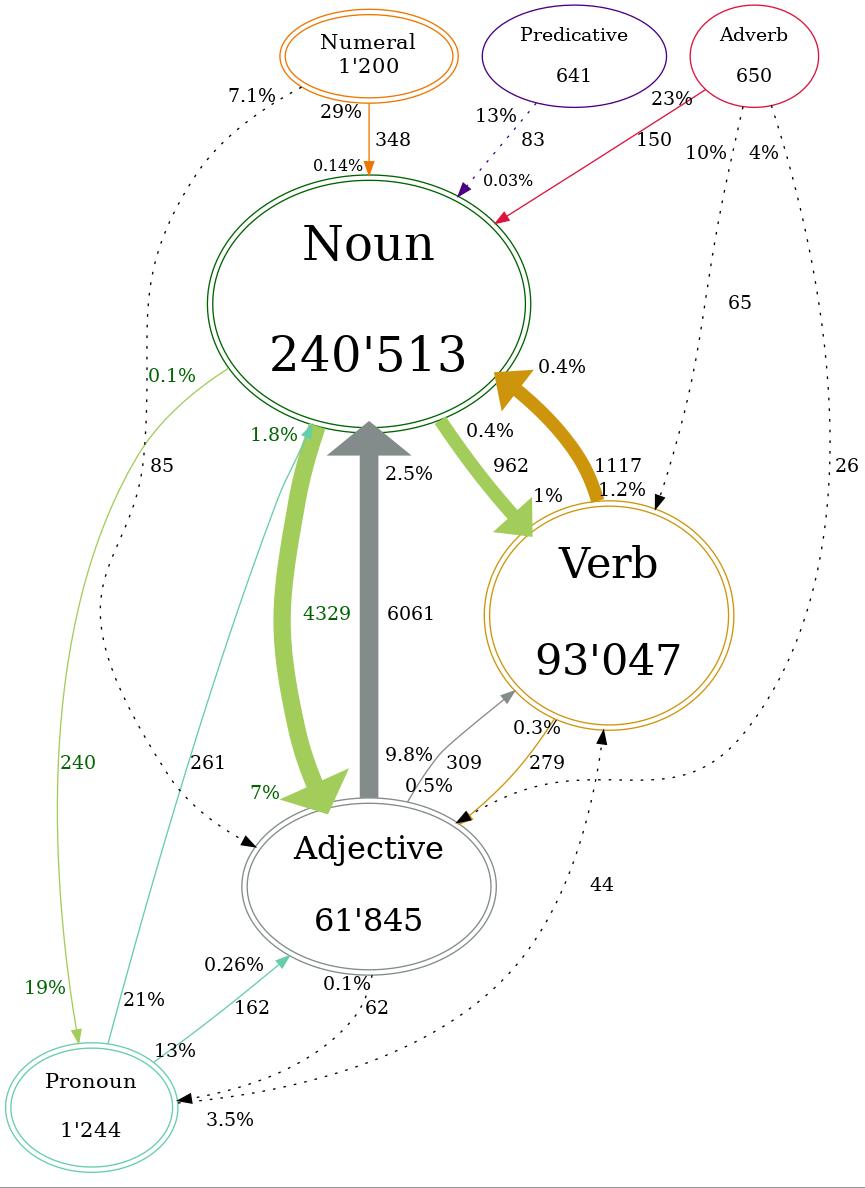}
  \caption{Veps language}
  \label{fig:pos-error-graph-vep}
\end{subfigure}%
\begin{subfigure}{.5\textwidth}
  \centering
\includegraphics[width=1.0\textwidth]{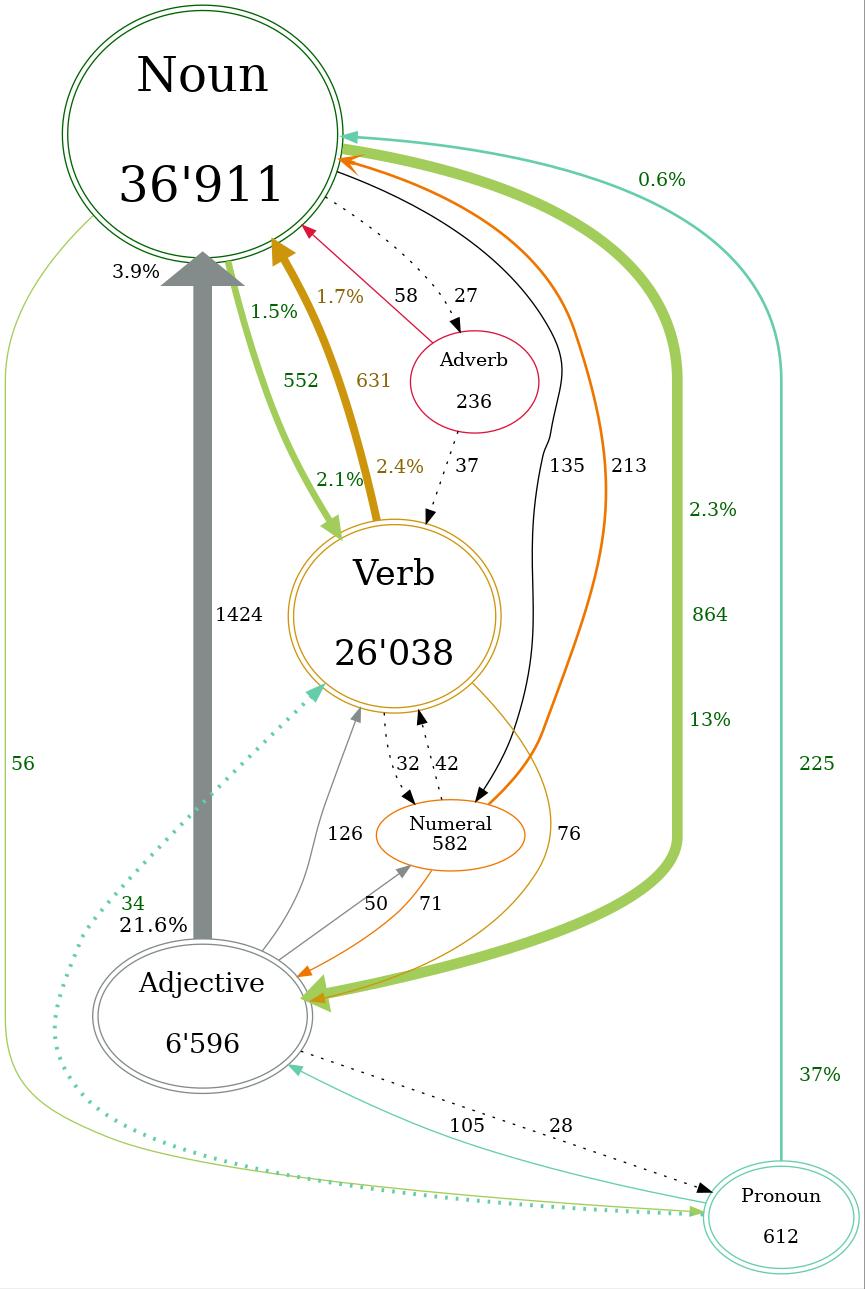}  \caption{Karelian Proper supradialect}
  \label{fig:pos-error-graph-krl}
\end{subfigure}
\caption{Part-of-speech error transition graph, which reflects the results of the POSGuess algorithm.}
\label{fig:pos-error-graph}
\end{figure}

In order to analyze the algorithm errors, the results of the part-of-speech algorithm POSGuess were visualized using the Graphviz program.
%
%
%
Part-of-speech error transition graphs were built for Veps language (Fig.~\ref{fig:pos-error-graph-vep}) 
and Karelian Proper supradialect (Fig.~\ref{fig:pos-error-graph-krl}).

Let us explain how these graphs were built.
%
For example, a thick grey vertical arrow connects adjective and noun (Fig.~\ref{fig:pos-error-graph-krl}), and this arrow has labels of 21.6\%, 1424 and 3.9\%.
%
This means that the POSGuess algorithm has erroneously identified \num{1424} Karelian adjectives as nouns.
%
This accounted for 21.6\% 
of all Karelian adjectives 
and 3.9\% of nouns.
%
This can be explained by the fact that 
the same lemma (in Veps and Karelian) can be both a noun and an adjective. 
Nouns and adjectives are inflected in the same form (paradigm).

The experiment showed that there are significantly more such lemmas (noun-adjective) for the Karelian language than for the Veps language (21.6\% versus 9.8\% in Fig.~\ref{fig:pos-error-graph}).
%
Although in absolute numbers Veps exceeds Karelian, namely: 6061 versus 1424 errors of this kind. 
This is because the Veps dictionary is larger in the VepKar corpus.

\section{Conclusion}

This research devoted to the low-resource Veps and Karelian languages. 

Algorithms for assigning part of speech tags to words and grammatical properties to words are presented in the article. 
These algorithms use our morphological dictionaries, where the lemma, part of speech and a set of grammatical features (gramset) are known for each word form. 

The algorithms are based on the analogy hypothesis that words with the same suffixes are likely to have the same inflectional models, the same part of speech and gramset.

The accuracy of these algorithms were evaluated and compared. 313 thousand Vepsian and 66 thousand Karelian words were used to verify the accuracy of these algorithms.
The special functions were designed to assess the quality of results of the developed algorithms.


71,091 ``word -- part of speech'' pairs for the Karelian Proper supradialect and 399,260 ``word -- part of speech'' pairs for the Veps language have been used in the experiments to evaluate algorithms.
86.77\% of Karelian words 
and 92.38\% of Vepsian words 
were assigned a correct part of speech. 

73,395 ``word -- gramset'' pairs for the Karelian Proper supradialect and 452,790 ``word -- gramset'' pairs for the Veps language have been used in the experiments to evaluate algorithms.
90.7\% of Karelian words 
and 
95.3\% of Vepsian words 
were assigned a correct gramset by our algorithm.

%
If you need only one correct answer, then all three of developed algorithms are not very useful.
%
%
But in our case, the task is to get an ordered list of the parts of speech and gramsets for a word and to offer this list to an expert.
Then the expert selects the correct part of speech and gramset from the list and assigns to the word. 
This is a semi-automatic tagging of the texts.
%
Thus, these algorithms are useful for our corpus.

%
%
%
\bibliographystyle{splncs04}
\bibliography{mybibliography}

\end{document}